# Active Deep Learning for Classification of Hyperspectral Images

Peng Liu, Hui Zhang, and Kie B. Eom

*Abstract*—Active deep learning classification of hyperspectral images is considered in this paper. Deep learning has achieved success in many applications, but good-quality labeled samples are needed to construct a deep learning network. It is expensive getting good labeled samples in hyperspectral images for remote sensing applications. An active learning algorithm based on a weighted incremental dictionary learning is proposed for such applications. The proposed algorithm selects training samples that maximize two selection criteria, namely representative and uncertainty. This algorithm trains a deep network efficiently by actively selecting training samples at each iteration. The proposed algorithm is applied for the classification of hyperspectral images, and compared with other classification algorithms employing active learning. It is shown that the proposed algorithm is efficient and effective in classifying hyperspectral images.

## I. Introduction

RECENTLY, a semisupervised learning method called deep learning [1] has been introduced for remote sensing data classification [2]–[4]. It can be considered as an extension of an artificial neural network, and is effective in classification of complex problems. However, training a deep network is quite expensive and requires a large number of training samples. The application of a deep network to hyperspectral image classification is not practical, because only a limited number of training samples are available and the dimension of the feature space is large.

Active learning is an iterative procedure of selecting the most informative examples from a subset of unlabeled samples. This choice is based on a ranking of scores that are computed from a model outcome. The chosen candidates are added to the training set, and the classifier is trained with the new training samples. The training done with actively selected samples is more efficient than the one done with randomly selected samples because it uses samples that are more suitable for training. Therefore, the active learning method can train a deep network faster and with fewer training samples than traditional semisupervised learning methods.

There are at least three different kinds of approaches in selecting new training samples for active learning. The first is based on the uncertainty of unlabeled samples, such as uncertain sampling [5], or query-by-committee [6]. The second is based on the influence on the model by the unlabeled samples such as length of gradients [7], or Fisher information ratio [8]. The third is based on the intrinsic distribution and structure of the unlabeled samples such as manifold learning [9], Kullback–Leibler (KL) divergence similarity [10], Gaussian similarity [11], and cluster [12]. There are also some mixed methods [13] that employ criteria in selecting new training samples for active learning. For example, the density-weighting method [13] employs both uncertainty and the distribution of the unlabeled samples. The methods using multiple metrics have the potential to achieve higher efficiency in active learning. Both uncertainty and distribution are utilized in selecting new training samples in the active learning algorithm proposed in this paper.

Active learning methods have been widely studied for remote sensing applications. Most of the research on active learning is combined with a special classifier or a special remote sensing application. Examples include a kernel-based method, an active learning method combined with a support vector machine (SVM) [14], logistic regression (LR) [15], and Gaussian process regression [16]. A survey for active learning in remote sensing before 2011 can be found in [17].

Although active learning has been applied to many applications in remote sensing, most of these approaches are closely connected with a specific type or a specific structure of the classifier. Examples are random sampling (RS), maximum uncertainty sampling (MUS) [18] and query-by-committee (QBC) [17]. RS is essentially a deep belief network (DBN) without fine tuning done by active learning. MUS and QBC are not applicable to DBN classifiers because both unsupervised feature learning and supervised fine tuning are employed in training of DBN. RS samples candidate data randomly, and the classification accuracy is usually low. However, it is fast, simple, and convenient. MUS queries the most uncertain instance by an active learner, and the entropy is used as an uncertainty measure. MUS has been applied to an SVM and LR. QBC trains committee members on the current labeled set. Different members of the committee represent different hypotheses of the classification problem. QBC selects candidate samples showing maximal disagreement between different members of the committee.

In this paper, we propose an active learning scheme where the information from both unsupervised and supervised stages is utilized. The proposed active learning is applied to a deep learning structure, and its efficacy in classifying remotely sensed

Manuscript received February 20, 2016; revised April 27, 2016, July 12, 2016, and July 31, 2016; accepted August 2, 2013. This work was supported in part by NSFC under Grant 41471368 and Grant 41571413 and in part by RADI Director Youth foundation. *(Corresponding author: Kie B. Eom.)*

P. Liu is with the Institute of Remote Sensing and Digital Earth, Chinese Academy of Sciences, Beijing 100094, China (e-mail: liupeng@radi.ac.cn).

H. Zhang is with the Laboratory of Brain and Cognition, National Institute of Mental Health, Bethesda, MD 20892 USA (e-mail: hui.zhang@nih.gov).

K. B. Eom is with the Department of Electrical and Computer Engineering, George Washington University, Washington, DC 20052 USA (e-mail: eom@gwu.edu).

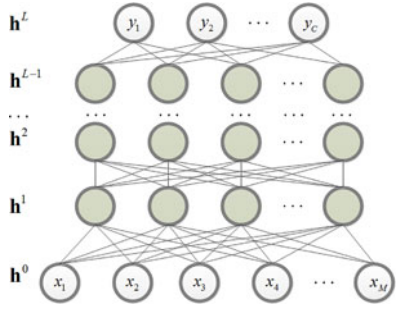

Fig. 1. DBN.

hyperspectral images is demonstrated in experiments. The paper is organized as follows: In Section II, the basic idea of the DBN is discussed, and it is shown that both unsupervised and supervised stages could provide useful information. In Section III, criteria for active learning is discussed, and two criteria, uncertainty and representativeness are proposed as criteria for active learning. In Section IV, an object function is constructed by combining representativeness and uncertainty of the samples, and the optimization algorithm for solving the new object function is presented. Experimental results are presented in Section V.

## II. DBN

Let $\mathcal{S} = \{1, \ldots, n, \ldots, N\}$ be a set of integers indexing $N$ pixels of a hyperspectral image, and let $\mathcal{L} = \{1, \ldots, c, \ldots, C\}$ be a set of integers indexing $C$ class labels. The image $X = \{\mathbf{x}_1, \ldots, \mathbf{x}_n, \ldots \mathbf{x}_N\}$ is a set composed of $N$ feature vectors, where $\mathbf{x}_n = \{x_n^1, \ldots, x_n^M\}$ is corresponding to the $n$th $M$-dimensional pixel. The label $Y = \{\mathbf{y}_1, \ldots, \mathbf{y}_n, \ldots \mathbf{y}_N\}$ is represented as a set composed of $N$ label vectors corresponding to $N$ pixels, where $\mathbf{y}_n = \{y_n^1, \ldots, y_n^C\}$ is $C$-dimensional label vector. The element $y_n^c$ in a label vector $\mathbf{y}_n$ represents the possibility that the pixel $\mathbf{x}_n$ belongs to the class $c$.

The labeling process can be considered as a mapping process from the image $X$ to the label $Y$, and it can be solved by a DBN. The DBN architecture used in this research is shown in Fig. 1. It is a fully interconnected belief network with one input layer $\mathbf{h}^0$, $L-1$ hidden layers $\mathbf{h}^1, \ldots, \mathbf{h}^{L-1}$, and an output layer $\mathbf{h}^L$. The input layer $\mathbf{h}^0$ has $M$ units corresponding to feature vector $\mathbf{x}_n$. The output layer $\mathbf{h}^L$ has $C$ units corresponding to label vector $\mathbf{y}_n$.

The DBN architecture transforms high-dimensional data into low-dimensional code using an adaptive and multilayered encoder network. One popular method for constructing a DBN deep architecture is the greedy layer-wise restricted Boltzmann machine (RBM). The RBM is a particular form of a log-linear Markov random field. Consider the layer $l$, it takes the output from the previous layer $\mathbf{h}^{l-1}$ as input (visible variable) and generates the output (hidden variable) for the next layer $\mathbf{h}^l$. For notational convenience, visible and hidden variables are denoted as $\mathbf{v}$ and $\mathbf{h}$, respectively.

There are no direct connections between hidden units in an RBM. The network assigns the following probability $p(\mathbf{v}, \mathbf{h})$ to every possible visible-hidden vector pair with the aforementioned energy function.

$$p(\mathbf{v}, \mathbf{h}) = \frac{1}{Z} e^{-E(\mathbf{v}, \mathbf{h})} \quad (1)$$

where the normalization term $Z$ is obtained by summing over all possible pairs of visible and hidden vectors

$$Z = \sum_{\mathbf{v}, \mathbf{h}} e^{-E(\mathbf{v}, \mathbf{h})}. \quad (2)$$

The probability $p(\mathbf{v})$ that the model assigns to a visible vector $\mathbf{v}$ is obtained by marginalizing over the space of hidden vectors

$$p(\mathbf{v}) = \frac{1}{Z} \sum_{\mathbf{h}} e^{-E(\mathbf{v}, \mathbf{h})}. \quad (3)$$

In RBMs, visible and hidden units are conditionally independent of each other. Therefore, conditional probabilities $p(\mathbf{h}|\mathbf{v})$ and $p(\mathbf{v}|\mathbf{h})$ can be written as

$$p(\mathbf{h}|\mathbf{v}) = \prod_i p(h_i|\mathbf{v}) \quad (4)$$

and

$$p(\mathbf{v}|\mathbf{h}) = \prod_j p(v_j|\mathbf{h}) \quad (5)$$

where the conditional activation probabilities are defined as follows [19]:

$$p(h_i = 1|\mathbf{v}) = f(W_i \mathbf{v} + b_i) \quad (6)$$
$$p(v_j = 1|\mathbf{h}) = f(W_j' \mathbf{h} + c_j) \quad (7)$$

where $W$ is the weight matrix, $\mathbf{b}$ and $\mathbf{c}$ are the offset vectors, and $f(\cdot)$ is the sigmoid function.

Considering the encoding and decoding of each layer, parameters $W$, $\mathbf{b}$, and $\mathbf{c}$ are related by the energy function $E(\mathbf{v}, \mathbf{h})$ of the RBM, and it is defined as

$$E(\mathbf{v}, \mathbf{h}) = -\mathbf{b}'\mathbf{v} - \mathbf{c}'\mathbf{h} - \mathbf{h}'W\mathbf{v}. \quad (8)$$

Furthermore, based on a series of derivations [19], [20], the log-likelihood gradient for the parameters of an RBM is obtained as

$$\frac{\partial \ln p(\mathbf{v})}{\partial \theta} = \sum_{\mathbf{h}} p(\mathbf{h}|\mathbf{v}) \frac{\partial E(\mathbf{v}, \mathbf{h})}{\partial \theta} - \sum_{\mathbf{v}, \mathbf{h}} p(\mathbf{v}, \mathbf{h}) \frac{\partial E(\mathbf{v}, \mathbf{h})}{\partial \theta} \quad (9)$$

where $\theta$ represents parameter $W$, $\mathbf{b}$, or $\mathbf{c}$.

In (9), the first term denotes an expectation with respect to the data distribution and the second tern denotes an expectation with respect to the distribution defined by the model. Because there are no direct connections between hidden units in an RBM, it is easy to get an unbiased sample of the first term. However, getting an unbiased sample of the second term is much more difficult. It can be done by starting at any random state of the visible unit and performing alternating Gibbs sampling for a very long time. A much faster learning procedure called constrastive divergence (CD) method was proposed in [19]. The CD performs Gibbs sampling, and uses a gradient descent procedure to update the increments of the parameters.

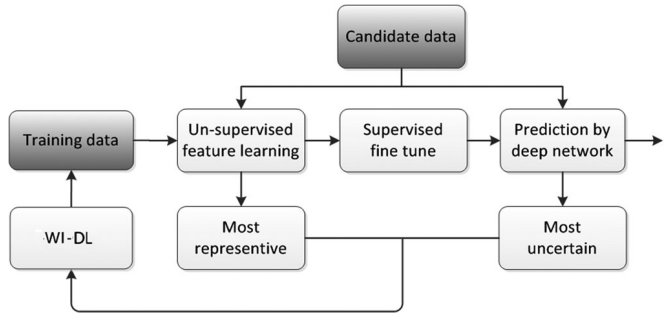

Fig. 2. Proposed active learning algorithm.

Once RBMs are stacked [1] and trained in a greedy manner, they form a DBN illustrated in Fig. 1. DBNs can be viewed as graphical models that learn to extract a deep hierarchical representation of the training data. A complete training of a DBN includes two stages: the unsupervised features learning stage and the supervised fine tuning stage. In the unsupervised features learning stage, RBMs learn one layer at a time by the CD method. This is an efficient greedy learning scheme. In the supervised fine tuning stage, all the initial RBM are stacked, and discriminative fine tuning is performed by a back propagation (BP) [21] algorithm.

In many applications, deep learning shows better performance than classical methods such as an SVM. However, deep learning needs to initialize more parameters than an SVM. A complete training of the DBN with randomly selected training samples requires a large number of training samples, despite the efficiency of the CD algorithm. Therefore, it is not practical to apply DBN to problems with limited size of training samples or with large feature spaces. How to select a training dataset becomes more important in training a deep network. To improve the training efficiency of the DBN, the algorithm proposed in this paper considers two stages in DBN training, unsupervised feature learning and supervised fine tuning. In the unsupervised learning stage, a DBN provides the condition for the estimation of the representativeness of data, while in the supervised learning stage, it provides the condition for the uncertainty estimation of data. In the proposed active learning method, the two metrics, representativeness and uncertainty, are integrated into an object function. The weighted incremental dictionary learning (WI-DL) algorithm illustrated in Fig. 2 is proposed to optimize the object function with two metrics. After the optimization, the samples are ranked and informative samples are selected. Experiments on hyperspectral images confirmed the effectiveness of the proposed algorithm.

In the following section, criteria for selecting training samples are discussed. It is also discussed how a deep network is trained by an active learning scheme.

## III. Criteria for Active Learning

In active learning, the training samples need to be selected by their importance. Criteria for selecting important samples for training in active learning have been considered in earlier research. An example of this research is a density-weighting method with the information density framework described by Settles and Craven [13]. It simultaneously considers underlying structure information of data and explicit class labels of the samples. It is also proposed that the most informative instances should not only be uncertain, but should also be representative of the underlying distribution [13]. Therefore, the most informative sample $\mathbf{x}^*$ is selected by maximizing the information density [13].

$$\mathbf{x}^* = \max_{\mathbf{x}} \left\{ \Phi(\mathbf{x}) \frac{1}{N} \sum_{n=1}^{N} S(\mathbf{x}, \mathbf{x}_n) \right\} \quad (10)$$

where $\mathbf{x}$ is a candidate sample, and $\mathbf{x}_n$ is an arbitrary sample in an unlabeled dataset $X$. The function $S(\mathbf{x}, \mathbf{x}_n)$ measures the similarity between the sample $\mathbf{x}$ and $\mathbf{x}_n$, and the function $\Phi(\mathbf{x})$ measures the uncertainty of the sample. The success of the information density approach depends on finding a good uncertainty function $\Phi(\mathbf{x})$ and a good similarity function $S(\mathbf{x}, \mathbf{x}_n)$. The information density framework has advantages in classification because it utilizes more information than other traditional active learning methods.

Inspired by the information density method [13], a similarity function $S(\mathbf{x}, \mathbf{x}_n)$ and an uncertainty function $\Phi(\mathbf{x})$ for a deep learning architectures are proposed in this paper. It is also important to combine the two criteria (similarity and uncertainty) together for searching informative samples for training. A new algorithm called the WI-DL is developed in the next section. This algorithm ranks the samples by their importance under the two criteria (similarity and uncertainty). The WI-DL algorithm selects informative samples that maximize the aforementioned two criteria as the new training data.

### A. Sparsity for Representativeness Estimation

A cosine similarity function was used as a similarity function in [13], and many other similarity measures, such as KL divergence similarity [10], Gaussian similarity [11], and local manifold similarity [9], have been used in active learning. In active learning, a good similarity measure is important in finding representative samples. In other words, a good representative model is needed for an efficient representation of the distribution and structure.

Recently, the sparse representation method, which represents a signal by a set of basis, has become popular in the machine learning field. A set of basis is also called a dictionary. Unlike decompositions using a predefined analytic basis (such as wavelet) and its variants, a signal can be represented using an overcomplete dictionary without analytic form. The basic assumption behind the dictionary learning approach is that the structure of complex incoherent characteristics can be extracted directly from the data rather than by using a mathematical description.

A set of atoms in a dictionary can characterize the entire dataset, as samples searched for active learning are representative. Furthermore, if a subset of samples is used as a dictionary with high efficiency, they are representative samples for an active learning problem.

In the deep architecture in Fig. 1, the output at the $l$th layer $H^l = \{\mathbf{h}_1^l, \ldots, \mathbf{h}_n^l, \ldots, \mathbf{h}_N^l\}$ can be considered as the projection of the $N$ input feature vectors $X = \{\mathbf{x}_1, \ldots, \mathbf{x}_n, \ldots \mathbf{x}_N\}$. Since the unsupervised coding stage of deep learning could be viewed as feature learning and dimension reduction, it is reasonable to select training samples based on the information of the feature data projected by the DBN. Therefore, the atoms can be selected from the last ($L$th) hidden layer output $H^L = \{\mathbf{h}_1^L, \ldots, \mathbf{h}_n^L, \ldots, \mathbf{h}_N^L\}$. Further, the output data $\mathbf{h}_n^L \in R^C$ can be described as $\mathbf{h}_n^L = D\alpha$, where $D \in R^{C \times p}$ is a dictionary with $p$ atoms [22]. Each atom in $D$ is normalized to a unit vector, and $\alpha \in R^p$ is the coefficient vector for sparse representation.

The dictionary is assumed redundant ($p > C$). The number of nonzero coefficients in the representation is denoted as $k = \|\alpha\|_0$, where $k$ is expected to be very small. It implies that the feature vector $\mathbf{h}_n^L$ can be viewed as a linear combination of a few columns from the dictionary $D \in R^{C \times p}$, which is also referred to as the set of atoms. For the convenience of notation, the dictionary learning problem can be stated without index $L$.

$$\min_{D,\alpha} \|\mathbf{h}_n - D\alpha\|_2^2 \quad \text{subject to} \quad \|\alpha\|_0 \leq k \quad (11)$$

where $\|\cdot\|_2$ is $L_2$ norm and $\|\cdot\|_0$ is $L_0$ norm.

Let $H = \{\mathbf{h}_1, \ldots, \mathbf{h}_N\}$ be the dataset with $N$ feature vectors, and let $A = \{\alpha_1, \ldots, \alpha_N\}$, be the set of corresponding coefficients, where $A \in R^{p \times N}$. Then, the dictionary learning problem can be written as

$$\min_{D,\alpha_1,\ldots,\alpha_N} \sum_{i=1}^{N} \|\mathbf{h}_i - D\alpha_i\|_2^2 + \lambda \sum_{i=1}^{N} \|\alpha_i\|_0 \quad (12)$$

where $\lambda$ is a regularization parameter.

Traditional methods for solving (12), such as K-singular value decomposition (K-SVD) [22], nonparametric Bayesian dictionary learning, etc., are not directly applicable to active learning with a deep network because uncertainty also needs be considered in addition to the representativeness.

### B. Information Entropy for Uncertainty Estimation

Uncertainty sampling [5] is a commonly used query framework for active learning. In this framework, an active learner queries the instances that are least certain. The entropy [5] is often used as an uncertainty measure.

$$\Phi(\mathbf{x}) = -\sum_{j=1}^{C} p(y_j|\mathbf{x}) \log(p(y_j|\mathbf{x})) \quad (13)$$

where $p(y_j|\mathbf{x})$ is the probability that the sample $\mathbf{x}$ belongs to the $j$th class. The information entropy $\Phi(\mathbf{x})$ of the sample $\mathbf{x}$ is based on the prediction of the current classifier.

In a deep network, an input $\mathbf{x}$ is projected to a hidden layer $\mathbf{h}$. The hidden layer $\mathbf{h}$ will obtain its label vector $\mathbf{y} = \{y_1, \ldots, y_j, \ldots, y_C\}$ after the classification prediction. The entropy $\Phi(\mathbf{h})$ of the hidden layer $\mathbf{h}$ is defined as

$$\Phi(\mathbf{h}) = -\sum_{j=1}^{C} p(y_j|\mathbf{h}) \log(p(y_j|\mathbf{h})) \quad (14)$$

where $p(y_j|\mathbf{h})$ is the probability that the sample $\mathbf{x}$ is mapped to the $\mathbf{h}$ and belongs to the $j$th class. The uncertainty function (14) is not directly related to the structures of the classifier, and is easy to implement. In the next section, the uncertainty measure is combined with sparse representation to develop an active learning algorithm.

## IV. Active Learning With Sparse Representation and Uncertainty

In the last section, we defined the representativeness measurement based on sparse representation and the uncertainty measurement based on entropy. In this section, we construct a new discriminate function that is used to search the most informative samples and to employ both representativeness and uncertainty of the samples.

For an active learning problem, when we select the most informative samples from the unlabeled dataset $X = \{\mathbf{x}_1, \ldots, \mathbf{x}_N\}$, the calculation is not directly performed in $X$ but in the corresponding projected data $H = \{\mathbf{h}_1, \ldots, \mathbf{h}_N\}$ at the output of the deep network. This is because $H$ will represent the features of input data more concisely and efficiently after the nonlinear dimension reduction. Therefore, the samples to be labeled are in $X$, but the searching process is done in the set $H$. There is a one-to-one relationship between $X$ and $H$. Once we find appropriate feature vectors in $H$, the corresponding samples in $X$ will be labeled and put into the training dataset.

The initial training data are projected by the deep network trained with the unsupervised learning stage that was explained in Section II, and the projected (output of the deep network) data are used as the initial dictionary $D$ (Usually atoms in $D$ are normalized to unit vector). For active learning, the dictionary $D = \{\mathbf{d}_1, \ldots, \mathbf{d}_n\}$ at the current iteration is appended with the new set of dictionary atoms $E = \{\mathbf{d}_{n+1}, \ldots, \mathbf{d}_{n+m}\} \subset H$, and the new dictionary for the next iteration is obtained. The selection of atoms is done similarly to a batch operation shown in (12), but the optimization is done incrementally. The object function for the incremental learning $J$ is given by

$$J(E, \beta_1, \ldots, \beta_N) = \sum_{i=1}^{N} \left\| \mathbf{h}_i - [DE] \begin{bmatrix} \alpha_i \\ \beta_i \end{bmatrix} \right\|_2^2 + \lambda \sum_{i=1}^{N} \|\beta_i\|_0 \quad (15)$$

where $\|\cdot\|_2$ is $L_2$ norm, $\|\cdot\|_0$ is $L_0$ norm, $D = [d_1, d_2, \ldots, d_n]$ is the dictionary from the previous iteration, and $E = [d_{n+1}, d_{n+2}, \ldots, d_{n+m}]$ is the new set of atoms to be appended to $D$ to form a new dictionary. The first term in (15) is the residual error after adding new training samples. The coefficient vectors $\alpha_i$ and $\beta_i$ are for the data $\mathbf{h}_i$, and are associated with dictionaries $D$ and $E$, respectively.

$$\alpha_i = \begin{bmatrix} \alpha_{i,1} \\ \vdots \\ \alpha_{i,n} \end{bmatrix} \quad \beta_i = \begin{bmatrix} \alpha_{i,n+1} \\ \vdots \\ \alpha_{i,n+m} \end{bmatrix}. \quad (16)$$

Equation (15) can be rewritten as

$$J(E, \beta_1, \ldots, \beta_N) = \sum_{i=1}^{N} \|\mathbf{r}_i - E\beta_i\|_2^2 + \lambda \sum_{i=1}^{N} \|\beta_i\|_0 \quad (17)$$

where $\mathbf{r}_i$ is the residual after the sparse coding of the data $\mathbf{h}_i$, with the dictionary $D$ from the previous iteration, is done.

$$\mathbf{r}_i = \mathbf{h}_i - D\alpha_i. \quad (18)$$

As discussed before, to combine the representativeness and uncertainty, we need to integrate the information of (17) and (15) to form a new object function

$$J(E, \beta_1, \ldots, \beta_N) = \sum_{i=1}^{N} \|\mathbf{r}_i - E\beta_i\|_2^2 + \lambda \sum_{i=1}^{N} \|\Gamma\beta_i\|_0 \quad (19)$$

where

$$\Gamma = \text{diag.} \, [\Phi(\mathbf{d}_{n+1}), \ldots, \Phi(\mathbf{d}_{n+m})] \quad (20)$$

and $\Phi(\mathbf{d}_i)$ is the entropy of the atom $\mathbf{d}_i$ (as well as $\mathbf{h}$) defined in (14). The more uncertain the atom $\mathbf{d}_i$ is, the larger $\Phi(\mathbf{d}_i)$ is. This means that, if one feature vector is mainly composed of very uncertain atoms, it will increase punishments to the object function. Furthermore, the large coefficients will exacerbate the influence of uncertainties from the atoms.

In (19), the problem of minimizing the object function is a joint optimization problem with respect to the set $E$ of new dictionary atoms and the set of coefficients $B = \{\beta_1, \ldots, \beta_N\}$. The cost function $J$ in (19) is not jointly convex, but is convex with respect to each of the two sets ($E$ and $B$) when the other one is fixed. There has been extensive research that focuses on how to find good dictionary atoms and how to represent the dataset sparsely. Usually, there are two stages in dictionary leaning algorithms: the sparse coding stage, which is searching the optimal solution for $B$, and the dictionary updating stage, which is to find the solution for $E$.

For the sparse coding stage, there are variety of atom selection schemes in many greedy-based algorithms such as orthogonal matching pursuit (OMP) [23], compressive sampling matched pursuit [24], and StageOMP [25]. However, none of the traditional sparse coding methods consider the uncertainty of atoms, because classification problems based on active learning were not considered in earlier research. While atoms for the new dictionary are selected from the unlabeled dataset $H$, all elements in $H = \{\mathbf{h}_1, \ldots, \mathbf{h}_N\}$ are potential candidates. It is difficult to traverse all the elements in a large dataset repeatedly when searching for atoms. To narrow it down, the entire dataset $H$ was sorted first by their uncertainty (from the most uncertain to the least uncertain). Let $\{\mathbf{h}_{\varsigma_1}, \ldots, \mathbf{h}_{\varsigma_N}\}$ be the samples of the set $H$ sorted by the uncertainty value $\Phi(\mathbf{h}_i)$, where $\mathbf{h}_{\varsigma_1}$ is the most uncertain sample. Assuming that $m$ samples are updated at each iteration of active learning, the top $m$ uncertain samples $\{\mathbf{h}_{\varsigma_1}, \ldots, \mathbf{h}_{\varsigma_m}\}$ from the sorted list are selected as the initial estimates of new atoms for the new dictionary.

$$E = \{\mathbf{d}_{n+1}, \ldots, \mathbf{d}_{n+m}\} = \{\mathbf{h}_{\varsigma_1}, \ldots, \mathbf{h}_{\varsigma_m}\}. \quad (21)$$

In the sparse coding stage of (19), $E$ is assumed to be known. In this paper, for the active learning problem, a weighted OMP is used for the sparse coding stage of (19). For a general OMP, each $\mathbf{r}_i$, $1 \leq i \leq N$, is coded by selecting atoms one by one that are the most similar to $\mathbf{r}_i$. This means that an arbitrary $\mathbf{r}_i$ is projected to each candidate atom $\mathbf{d}_\eta$, $n+1 \leq \eta \leq n+m$, and the best match is selected by using

$$\mathbf{d} = \underset{\eta=n+1,\ldots,n+m}{\arg\max} \; |\mathbf{r}_i \cdot \mathbf{d}_\eta|. \quad (22)$$

However, the uncertainty $\Phi(\mathbf{d}_\eta)$ should be considered for minimizing (19), and the selection of the candidate atom is done by finding the match that gives maximum weighted projection value.

$$\mathbf{d} = \underset{\eta=n+1,\ldots,n+m}{\arg\max} \; \{\Phi(\mathbf{d}_\eta)|\mathbf{r}_i \cdot \mathbf{d}_\eta|\}. \quad (23)$$

An algorithm that maximizes the aforementioned equation (called weighted OMP) was developed, and is described in detail in Algorithm 2.

For the dictionary updating stage, the atom updating algorithm will be different from general dictionary learning methods because the atoms are the samples selected directly from the candidate dataset. The initial set of atoms $E = \{\mathbf{d}_{n+1}, \ldots, \mathbf{d}_{n+m}\}$ contains the $m$ most uncertain samples, but they are not good enough to be used for the new dictionary and should be updated one by one. Using the idea of information density in (10), both representativeness and uncertainty are considered. For the convenience of notation, the symbol $\mathbf{d}$ (unit vector) is used without an index. Representativeness is measured by the square of the inner product between the atom $\mathbf{d}$ and the vector $\mathbf{r}_i$, $1 \leq i \leq N$, as they are similar if this measure is large.

$$(\mathbf{d} \cdot \mathbf{r}_i)^2 = (\mathbf{d}^T \mathbf{r}_i)^2. \quad (24)$$

Therefore, an atom $\mathbf{d}$ is very representative if it is similar to all vectors in the set $R = \{\mathbf{r}_1 \ldots \mathbf{r}_N\}$. Therefore

$$\hat{\mathbf{d}} = \max_{\mathbf{d}} \sum_{i=1}^{N} (\mathbf{d}^T \mathbf{r}_i)^2. \quad (25)$$

However, the importance for each term $(\mathbf{d}^T \mathbf{r}_i)$ is different because of the difference in uncertainty. As the uncertainty of the sample increases, it contributes more to the sum. As a result, the uncertainty measure $\Phi(\mathbf{r}_i)$ is introduced into the object function

$$\hat{\mathbf{d}} = \max_{\mathbf{d}} \sum_{i=1}^{N} \Phi(\mathbf{r}_i)(\mathbf{d}^T \mathbf{r}_i)^2. \quad (26)$$

By changing the form of $\sum_{i=1}^{N} \Phi(\mathbf{r}_i)(\mathbf{d}^T \mathbf{r}_i)^2$, we get

$$\sum_{i=1}^{N} \Phi(\mathbf{r}_i)(\mathbf{d}^T \mathbf{r}_i)^2 = \sum_{i=1}^{N} \mathbf{d}^T \mathbf{r}_i \Phi(\mathbf{r}_i) \mathbf{r}_i^T \mathbf{d}$$

$$= \mathbf{d}^T \left( \sum_{i=1}^{N} \mathbf{r}_i \Phi(\mathbf{r}_i) \mathbf{r}_i^T \right) \mathbf{d}. \quad (27)$$

As addressed before, atom $\mathbf{d}$ is a unit vector, and $\mathbf{d}^T \mathbf{d} = 1$. Introducing the Lagrange multiplier $\xi$, the object function

**Algorithm 1** The Weighted OMP

**Input**: The target dataset $R = \{\mathbf{r}_1, \ldots, \mathbf{r}_N\}$, the set of atoms to be appended to the dictionary $E = \{\mathbf{d}_{n+1}, \ldots, \mathbf{d}_{n+m}\}$ and the sparsity level of the signal $k$.
**Output**: A coefficient matrix $B = \{\beta_1, \ldots, \beta_N\}$.
1) $B = \emptyset$.
**for** $i := 1$ **to** $N$ **do**
  2) Take $\mathbf{r}_i$ from $R$, and let $\hat{\mathbf{r}}_0 = \mathbf{r}_i$.
  3) Set index set $\Lambda = \emptyset$ and temporal dictionary $\hat{D} = \emptyset$.
  **for** $j := 1$ **to** $k$ **do**
    4) Find an index $\eta_j$ that solves the optimization problem
$$\eta_j = \arg\max_{\eta = n+1, \ldots, n+m} \Phi(\mathbf{d}_\eta)|\hat{\mathbf{r}}_{j-1} \cdot \mathbf{d}_\eta|.$$
    5) Augment the index set and the matrix of chosen atoms
$$\Lambda = \Lambda \cup \{\eta_j\}$$
$$\hat{D} = \hat{D} \cup \{\mathbf{d}_{\eta_j}\}.$$
    6) Solve a least squares problem to obtain a newly estimated sparse coefficients $\hat{\beta}_j$
$$\hat{\beta}_j = \arg\min_\beta \|\hat{\mathbf{r}}_{j-1} - \hat{D}\beta\|_2$$
    7) Calculate the new approximation of the current data and the new residual
$$\mathbf{b}_j = \hat{D}\hat{\beta}_j$$
$$\hat{\mathbf{r}}_j = \hat{\mathbf{r}}_{j-1} - \mathbf{b}_j$$
  **end for**
  8) $\beta_i = \hat{\beta}_j$, and $B = B \cup \beta_i$
**end for**

**Algorithm 2** The Weighted Incremental Dictionary Learning

**Input**: The current residual set $R = \{\mathbf{r}_1, \ldots, \mathbf{r}_N\}$. Candidate dataset $\Omega = \{\mathbf{h}_1, \ldots, \mathbf{h}_r\}$. The number of dictionary samples updated $m$.
**Output**: The new set atoms to be appended to the dictionary $E = \{\mathbf{d}_{n+1}, \ldots, \mathbf{d}_{n+m}\}$, and the new residual set $R'$.
1) Calculate uncertainty functions $\{\Phi(\mathbf{h}_1), \ldots, \Phi(\mathbf{h}_r)\}$ on the dataset $\Omega$ by (14), and then, sort them from the most uncertain to the least uncertain samples.
$$\Theta = \text{sort}(\Phi(\mathbf{h}_1), \ldots, \Phi(\mathbf{h}_r)).$$
2) From the set $\Theta$, select the top $m$ samples as candidates for the initial dictionary $E = \{\mathbf{d}_{n+1}, \ldots, \mathbf{d}_{n+m}\}$.
3) With $R$ and $E$, the parameter set $B$ is obtained by calling Algorithm 1.
**for** $a := 1$ **to** $m$ **do**
  4) For atom $\mathbf{d}_{n+a} \in E$ construct $\hat{R}_{t+1}\hat{W}\hat{R}_{t+1}^T$ by (32) and (33).
  5) Perform SVD and get $[U, \Lambda, V] = \text{svd}(\hat{R}_{t+1}\hat{W}\hat{R}_{t+1}^T)$.
  6) Take out the first column of $U = \{\hat{\mathbf{d}}_1, \ldots, \hat{\mathbf{d}}_B\}$ and find $\mathbf{h}^*$ in $\Omega$ by (35).
  7) $\mathbf{d}_{n+a} = \mathbf{h}^*$, update dictionary $E$ by $\mathbf{d}_{n+a}$, and update $B$ by (36).
  8) delete $\mathbf{h}^*$ from $\Omega$.
**end for**
9) Compute the residual set for the next iteration:
$$R' = R - EB$$

becomes

$$J_1(\mathbf{d}) = \mathbf{d}^T \left(\sum_{i=1}^N \mathbf{r}_i \Phi(\mathbf{r}_i) \mathbf{r}_i^T\right) \mathbf{d} - \xi(\mathbf{d}^T \mathbf{d} - 1). \quad (28)$$

By setting the partial derivative of $J_1$ with respect to $\mathbf{d}$ to zero, we get

$$\frac{\partial J_1(\mathbf{d})}{\partial \mathbf{d}} = \left(\sum_{i=1}^N \mathbf{r}_i \Phi(\mathbf{r}_i) \mathbf{r}_i^T\right) \mathbf{d} - \xi \mathbf{d} = 0. \quad (29)$$

This can be rewritten in the following form.

$$\xi \mathbf{d} = \left(\sum_{i=1}^N \mathbf{r}_i \Phi(\mathbf{r}_i) \mathbf{r}_i^T\right) \mathbf{d}. \quad (30)$$

The atom $\mathbf{d}$ is the eigenvector of the symmetric matrix $(\sum_{i=1}^N \mathbf{r}_i \Phi(\mathbf{r}_i) \mathbf{r}_i^T)$, and $\xi$ will be the eigenvalue of $(\sum_{i=1}^N \mathbf{r}_i \Phi(\mathbf{r}_i) \mathbf{r}_i^T)$. To solve this problem, a singular value decomposition (SVD) can be applied.

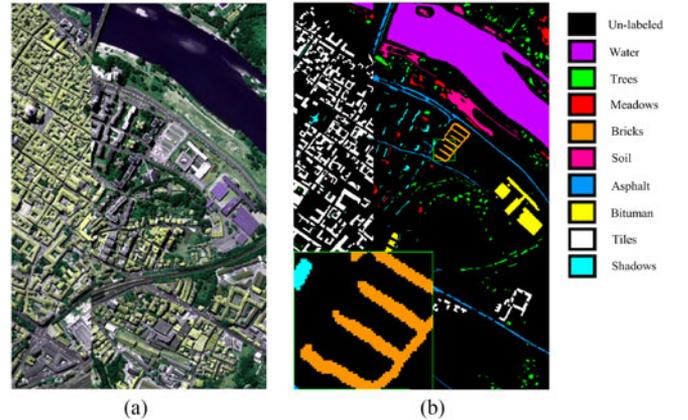

Fig. 3. Three-channel color composite image and ground truth of PaviaC data. (a) Image, (b) Ground truth.

However, it is unnecessary that all vectors in the set $R = \{\mathbf{r}_1, \ldots, \mathbf{r}_N\}$ need to be involved in updating one atom $\mathbf{d}$. Only a few coefficients in $B$ are nonzero after performing the weighted OMP of $R$, since it is sparse coding. As a result, if we want to update an arbitrary atom $\mathbf{d}_{n+a}$ ($1 \le a \le m$) by searching a new $\mathbf{d}$, only samples with nonzero coefficients should be considered in the search. Therefore, before updating $\mathbf{d}_{n+a}$, a subset $\{\mathbf{r}_{\zeta_1}, \ldots, \mathbf{r}_{\zeta_f}\} \subset R$ of data with nonzero coefficients (from $B$) for atom $\mathbf{d}_{n+a}$ is selected. For every $\mathbf{r}_{\zeta_f}$, it is encoded by $\{\mathbf{d}_{n+1},\ldots,\mathbf{d}_{n+a},\mathbf{d}_{n+a+1},\ldots,\mathbf{d}_{n+m}\}$ considering the influence without $\mathbf{d}_{n+a}$. Then, the remaining residual $\hat{\mathbf{r}}_{\zeta_f}$ is defined as

$$\hat{\mathbf{r}}_{\zeta_f} = \mathbf{r}_{\zeta_f} - \sum_{i=1}^{a-1} \mathbf{d}_{n+i}\beta_{\mathbf{d}_{n+i}} - \sum_{i=a+1}^{m} \mathbf{d}_{n+i}\beta_{\mathbf{d}_{n+i}} \quad (31)$$

where $\mathbf{d}_{n+i}$ is the atom, and $\beta_{\mathbf{d}_{n+i}}$ is corresponding coefficients from $B$ by the most recent weighted OMP. Therefore, we define

$$\hat{R} = \{\hat{\mathbf{r}}_{\zeta_1},\ldots,\hat{\mathbf{r}}_{\zeta_f}\}, \text{ and} \quad (32)$$

$$\hat{W} = \text{diag}.[\Phi(\hat{\mathbf{r}}_{\zeta_1}),\ldots,\Phi(\hat{\mathbf{r}}_{\zeta_f})]. \quad (33)$$

TABLE I
PAVIAC DATASETS

| Class | Label | Total | Dataset 1 | | | Dataset 2 | | | Dataset 3 | | |
|---|---|---|---|---|---|---|---|---|---|---|---|
| | | | Train (3%) | Candidate (20%) | Test (77%) | Train (5%) | Candidate (20%) | Test (75%) | Train (7%) | Candidate (20%) | Test (73%) |
| Water | 1 | 65971 | 1979 | 13194 | 50798 | 3299 | 13194 | 49478 | 4618 | 13194 | 48159 |
| Trees | 2 | 7598 | 228 | 1520 | 5850 | 380 | 1520 | 5698 | 532 | 1520 | 5546 |
| Meadows | 3 | 3090 | 93 | 618 | 2379 | 155 | 618 | 2317 | 216 | 618 | 2256 |
| Bricks | 4 | 2685 | 81 | 537 | 2067 | 134 | 537 | 2014 | 188 | 537 | 1960 |
| Soil | 5 | 6584 | 198 | 1317 | 5069 | 329 | 1317 | 4938 | 461 | 1317 | 4806 |
| Asphalt | 6 | 9248 | 277 | 1850 | 7121 | 462 | 1850 | 6936 | 647 | 1850 | 6751 |
| Bitumen | 7 | 7287 | 219 | 1457 | 5611 | 364 | 1457 | 5466 | 510 | 1457 | 5320 |
| Tiles | 8 | 42826 | 1285 | 8565 | 32976 | 2141 | 8565 | 32120 | 2998 | 8565 | 31263 |
| Shadows | 9 | 2863 | 86 | 573 | 2204 | 573 | 143 | 573 | 200 | 573 | 2090 |
| Total | - | 148152 | 4446 | 29631 | 114075 | 7407 | 29631 | 111114 | 10370 | 29631 | 108151 |

The term $(\sum_{i=1}^{N}\mathbf{r}_i\Phi(\mathbf{r}_i)\mathbf{r}_i^T)$ in (30) can be rewritten as $\hat{R}\hat{W}\hat{R}^T$. Similarly, we have $\xi\mathbf{d} = (\hat{R}\hat{W}\hat{R}^T)\mathbf{d}$. Performing SVD decomposition leads to

$$[U, \Lambda, V] = \text{svd}(\hat{R}\hat{W}\hat{R}^T) \quad (34)$$

where $U = \{\hat{\mathbf{d}}_1,\ldots,\hat{\mathbf{d}}_B\}$ are eigenvectors for matrix $\hat{R}\hat{W}\hat{R}^T$. Let $\hat{\mathbf{d}}_1$ be the eigenvector corresponding to the largest eigenvalue. A sample from $H$ is selected so that the projection to $\hat{\mathbf{d}}_1$ weighted with uncertainty is the largest. This selection process can be written as

$$\mathbf{h}^* = \max_{\mathbf{h}} \Phi(\mathbf{h})\left(\hat{\mathbf{d}}_1^T\mathbf{h}\right)^2. \quad (35)$$

Now, $\mathbf{h}^*$ is used as the new atom of $\mathbf{d}_{n+a}$ and $\mathbf{h}^*$ is put back into $E$ in the position of $\mathbf{d}_{n+a}$. At the same time, $B$ also needs to be updated. The coefficients corresponding to $\mathbf{d}_{n+a}$ in $B$ are calculated by

$$\nu_{n+a} = (\mathbf{d}_{n+a}^T\mathbf{d}_{n+a})^{-1}\mathbf{d}_{n+a}^T\hat{R}. \quad (36)$$

Then, $\nu_{n+a}$ is put into its corresponding position in $B$ in place of the old value. Atoms in $E = \{\mathbf{d}_{n+1},\ldots,\mathbf{d}_{n+m}\}$ will be updated one by one. Another problem is that, in (21), the initial value of $E$ is selected form $H$. Actually, if $H$ is too large, a subset $\Omega \subset H$ can be randomly selected from $H$ and used as a candidate dataset for the potential atoms in active learning. The complete algorithm for searching $\mathbf{d}_{n+1},\ldots,\mathbf{d}_{n+m}$ is given in Algorithm 2.

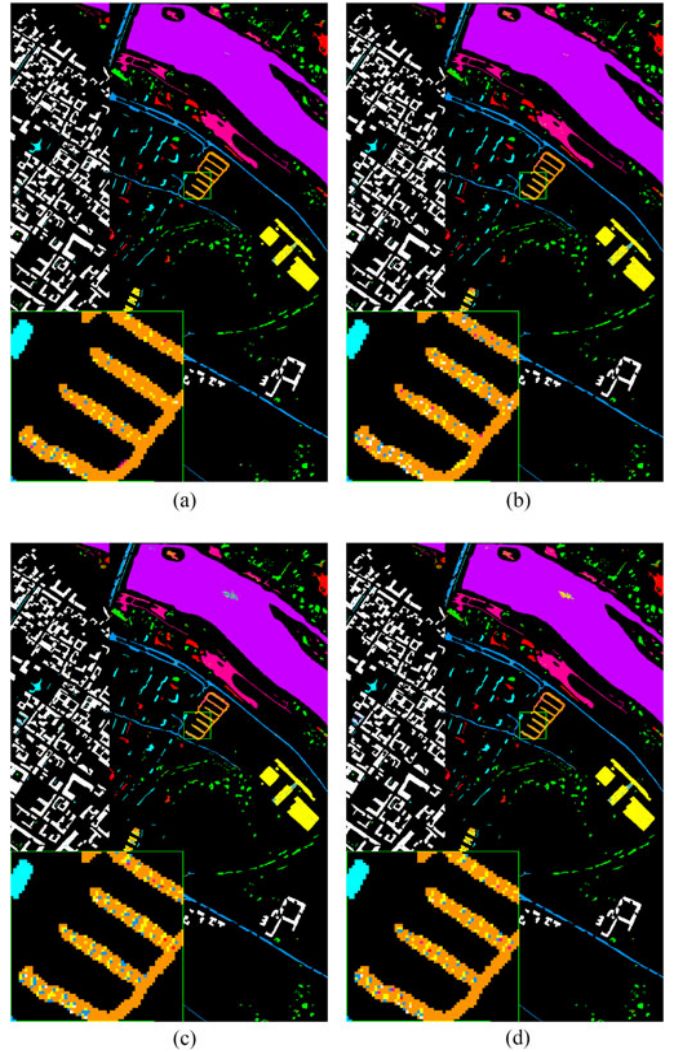

Fig. 4. Classification results of different active learning methods. The proposed WI-DL result shows better results compared with the ground truth in Fig. 3(b). The accuracies for WI-DL, MUS, RS, and QBC are 97.2%, 94.5%, 94.0%, and 95.5%, respectively. (a) WI-DL, (b) MUS, (c) RS, (d) QBC.

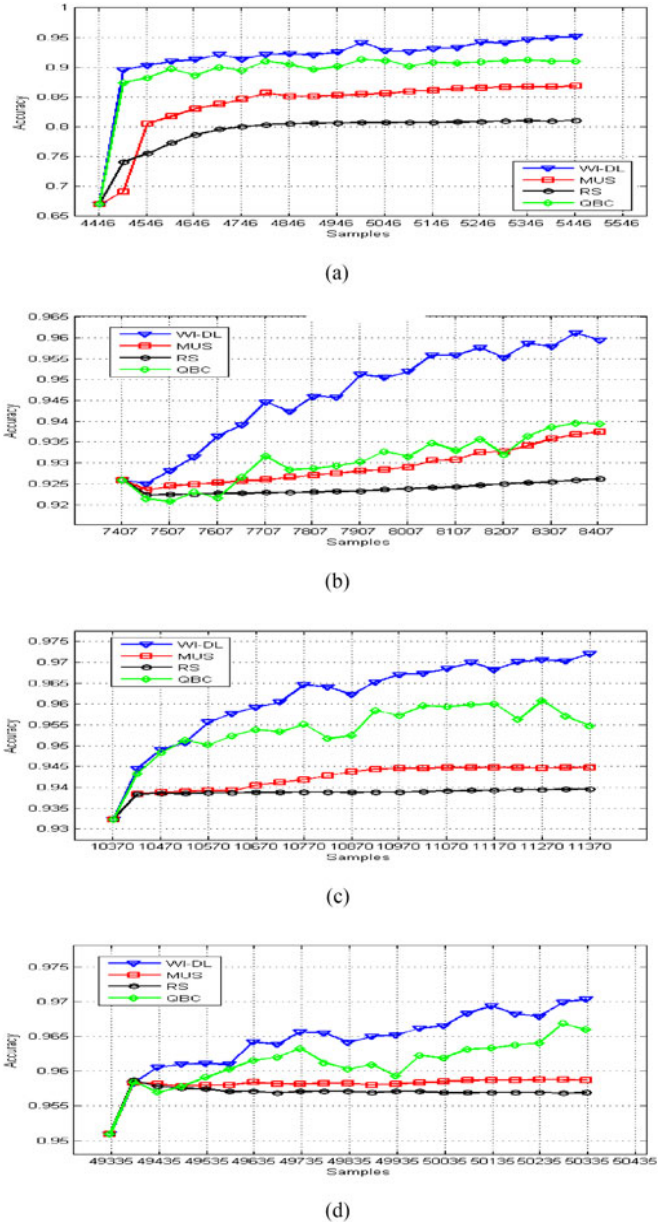

Fig. 5. Classification accuracy of different methods on different datasets. (a) Dataset 1. (b) Dataset 2. (c) Dataset 3. (d) Cross validation of 12 runs.

TABLE II
COMPUTATION TIME (IN SECONDS) FOR PAVIAC DATASETS

| | WI-DL | MUS | RS | QBC |
|---|---|---|---|---|
| Dataset 1 | 861 | 417 | 391 | 970 |
| Dataset 2 | 1622 | 1171 | 1165 | 2219 |
| Dataset 3 | 2521 | 1656 | 1479 | 3117 |

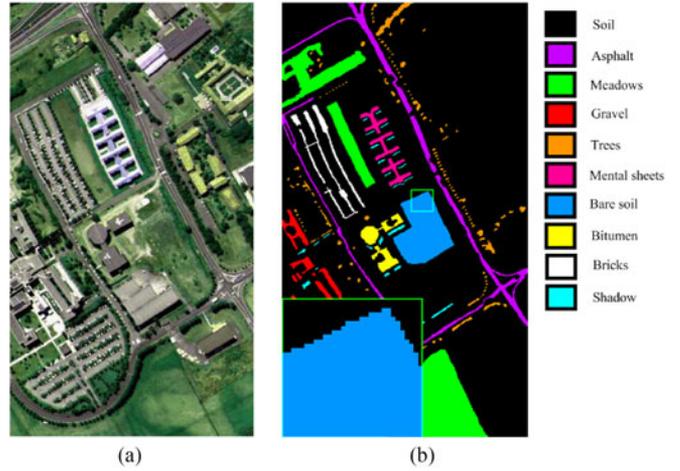

Fig. 6. Three-channel color composite image and ground truth of PaviaU data. (a) Image and (b) Ground truth.

is small, and the speedy unsupervised learning only brings the DBN to an initial configuration. The complete training of a DBN requires a large training set, and it is computationally expensive. The additional dictionary atoms selected at each iteration of active learning trains the DBN classifier much more efficiently than randomly selected training samples as the additional dictionary atoms are the $m$ best (in terms of representativeness and uncertainty) samples for training. In the experiment, $m$ atoms are added to the dictionary at each iteration. After each iteration of active learning, the DBN classifier is trained with the training set that is updated with additional $m$ atoms. This process is repeated until the DBN classifier is completely trained, and details are summarized in the experimental results.

When compared with the traditional dictionary learning algorithm, the proposed WI-DL has its own distinctive characteristics. First, it has a different object function from the traditional one because of the introduced weight parameters $\Phi(\mathbf{h})$. Furthermore, the uncertainty metric is considered in samples (as well as atoms) sorting and selecting procedures. Second, the traditional dictionary learning is applied to all signal or vector data, while WI-DL is only applied to the current residual data, because active learning is an incremental learning problem. Overall, WI-DL considered both the representativeness and the uncertainty while selecting samples to be atoms.

The active learning algorithm presented previously has been applied to improve the DBN classifier. The size of the training set

## V. EXPERIMENTS AND RESULTS

To validate the proposed method, three hyperspectral datasets, PaviaC, PaviaU, and Botswana are used in the experiment. The proposed algorithm, WI-DL, is compared on the test datasets with three other algorithms, namely RS, MUS [18], and QBC [17].

The DBNs used in this paper have four hidden layers. Computational efficiency is considered in selection of number of layers. The initial weights for DBNs are randomly selected between 0 and 1. Each layer of DBNs is based on an RBM. Once RBMs are stacked [1] and trained in a greedy manner, they form a DBN architecture illustrated in Fig. 1. In the unsupervised features learning stage, RBMs learn one layer at a time by the CD method in [15]. In the supervised fine-tuning stage, a BP [17] algorithm is applied.

TABLE III
PAVIAU DATASETS

| Class | Label | Total | Dataset 1 | | | Dataset 2 | | | Dataset 3 | | |
|---|---|---|---|---|---|---|---|---|---|---|---|
| | | | Train (10%) | Candidate (20%) | Test (70%) | Train (20%) | Candidate (20%) | Test (60%) | Train (30%) | Candidate (20%) | Test (50%) |
| Asphalt | 1 | 6631 | 663 | 1326 | 4642 | 1326 | 1326 | 3979 | 1989 | 1326 | 3316 |
| Meadows | 2 | 18649 | 1865 | 3730 | 13054 | 3730 | 3730 | 11189 | 5595 | 3730 | 9324 |
| Gravel | 3 | 2099 | 210 | 420 | 1469 | 420 | 420 | 1259 | 630 | 420 | 1049 |
| Trees | 4 | 3064 | 306 | 613 | 2145 | 613 | 613 | 1838 | 919 | 613 | 1532 |
| MentalSheets | 5 | 1345 | 135 | 269 | 941 | 269 | 269 | 807 | 404 | 269 | 672 |
| BareSoil | 6 | 5029 | 503 | 1006 | 3520 | 1006 | 1006 | 3017 | 1509 | 1006 | 2514 |
| Bitumen | 7 | 1330 | 133 | 266 | 931 | 266 | 266 | 798 | 399 | 266 | 665 |
| Bricks | 8 | 3682 | 368 | 736 | 2578 | 736 | 736 | 2210 | 1105 | 736 | 1841 |
| Shadows | 9 | 947 | 95 | 189 | 663 | 189 | 189 | 569 | 284 | 189 | 474 |
| Total | - | 42776 | 4278 | 8555 | 29943 | 8555 | 8555 | 25666 | 12834 | 8555 | 21387 |

## A. Experiment 1

The first experiment is done with the Pavia Center (PaviaC) dataset. It was acquired by a Reflective Optics System Imaging Spectrometer (ROSIS) sensor, and has been widely used in earlier research. The number of bands in the original dataset is 115, and spatial resolution is 1.3 m. It covers a spectral range from 0.43 to 0.86 $\mu$m with 115 hyperspectral bands. From the original PaviaC dataset, 102 bands are selected by removing low signal-to-noise ratio (SNR) bands. Test images are segmented from the dataset without low SNR bands, and the size of each test image is 1096 × 715 pixels. Fig. 3(a) shows a test image in false color, and Fig. 3(b) shows the ground truth with the detailed view at the lower left corner. It shows ten classes in different colors, and names of class labels are shown on the right side of the figure.

Nine classes of interest (Water, Trees, Meadows, Bricks, Soil, Asphalt, Bitumen, Tiles, and Shadows) have been selected for the labeled dataset. Four algorithms (WI-DL, RS, MUS, and QBC) are applied to three sets of data constructed from the ground-truth data. Each set contains three classes of randomly selected data, training, candidates, and testing data, of different percentages. Table I shows the number of samples for each class of the dataset. The class name and the class number are given in the first and second columns, while the third column shows the total number of samples, and the rest of columns show numbers of samples in training, candidates, and test sets.

A DBN having four hidden layers, 102 input nodes corresponding to 102 hyperspectral bands, and nine output nodes corresponding to nine classes of interest, is created. The training data are used to configure the parameters of the DBN classifier for the preparation of active learning. Then, the active learning algorithm presented in Section IV is applied to fine-tune the DBN classifier by actively selecting atoms from the candidate set. The training data used for the initial configuration of DBN is used as the initial dictionary, and 50 samples are actively selected from the candidates set at each iteration of active learning. The newly selected dictionary samples are labeled and added to the existing dictionary, and the DBN classifier is fine tuned with the updated dictionary. A total of 20 iterations are performed,

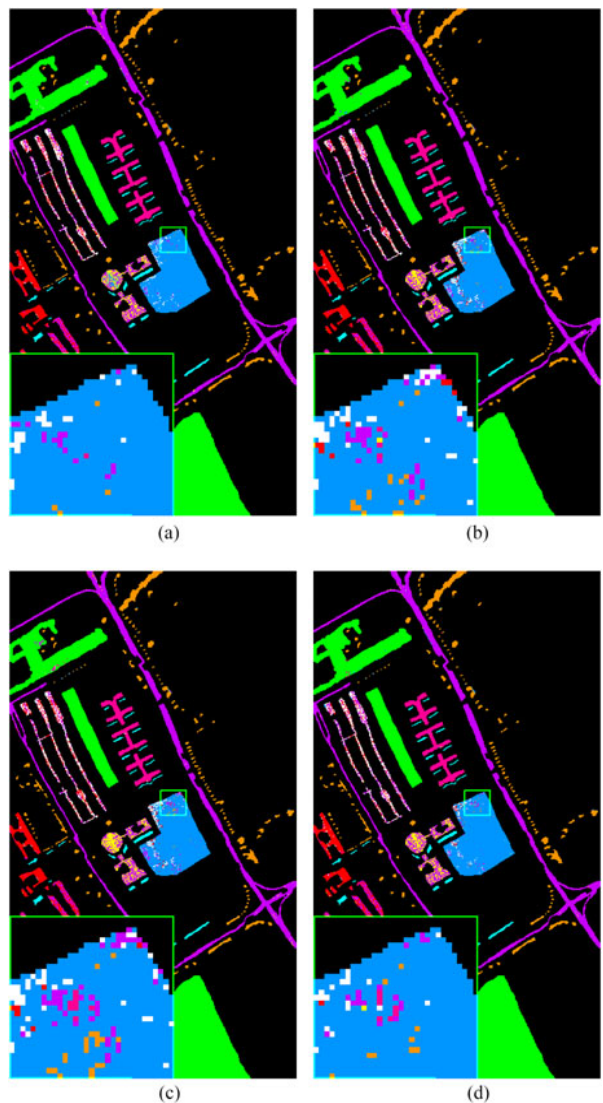

Fig. 7. Maps of classifications of different active learning methods. The accuracies for WI-DL, MUS, RS, and QBC are 92.4%, 78.3%, 72.2%, and 88.5%, respectively. (a) WI-DL, (b) MUS, (c) RS, (d) QBC.

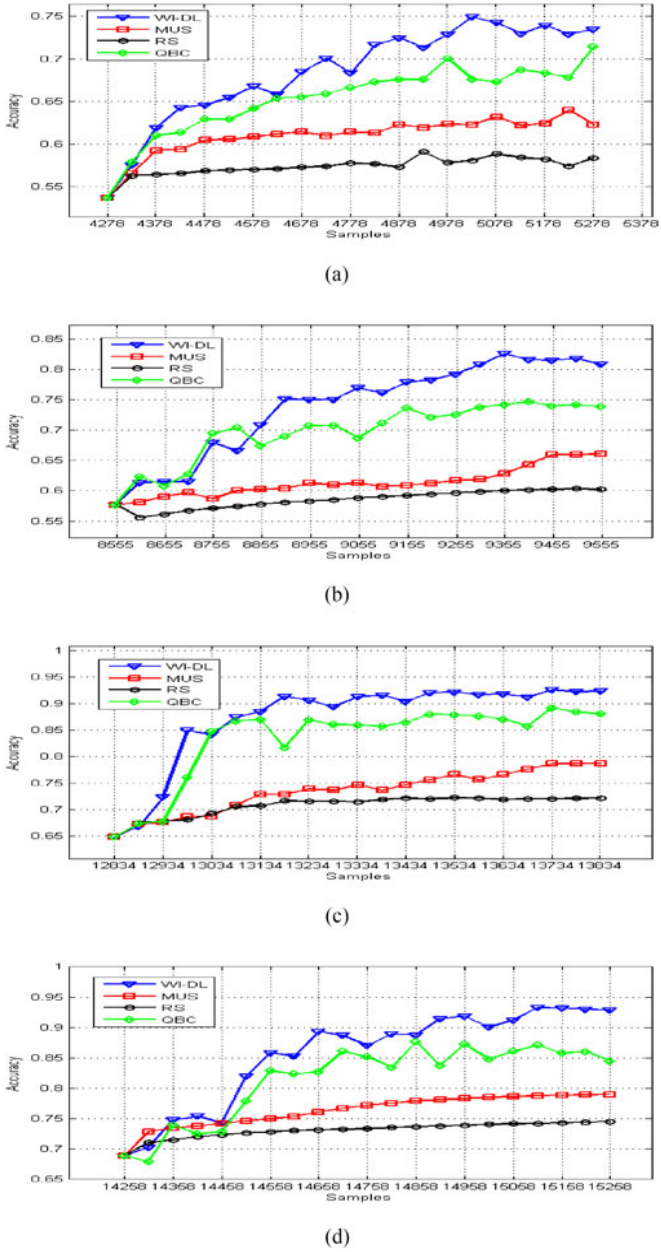

Fig. 8. Classification accuracy of different methods on different datasets. (a) Dataset 1. (b) Dataset 2. (c) Dataset 3. (d) Cross validation of 12 runs.

TABLE IV
COMPUTATION TIME (IN SECONDS) FOR PAVIAU DATASETS

|  | WI-DL | MUS | RS | QBC |
|---|---|---|---|---|
| Dataset 1 | 696 | 396 | 394 | 735 |
| Dataset 2 | 1921 | 945 | 936 | 2312 |
| Dataset 3 | 2309 | 1293 | 1184 | 2725 |

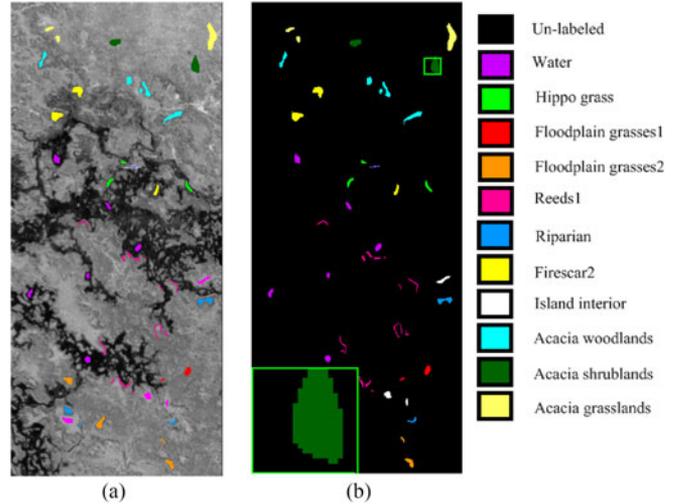

Fig. 9. Gray image and ground truth of Botswana data. (a) Image and (b) Grand truth.

and additional 1000 atoms are added to the dictionary in the active learning stage.

The test dataset was used to test the performance of the fine-tuned DBN classifier. Fig. 4(a) shows the classification result obtained by the proposed WI-DL algorithm, and the boxed area at the center is enlarged in the lower left corner to show classification details. In comparison with the image in Fig. 3(b), it can be seen that the classification result matches reasonably well with the ground truth. Three other active-learning classification algorithms, namely RS, MUS, and QBC methods, are applied to the same test dataset to compare classification performances. The classification results are shown in Fig. 4. The WI-DL result in Fig. 4(a) shows that more samples are correctly classified than compared to the results of other algorithms. It can be observed that the WI-DL algorithm performs better than other approaches, especially in the enlarged area in the lower left corner.

To demonstrate the effectiveness of active learning, the classification accuracy is measured with varying amount of training samples. The result of experiments are shown in Fig. 5. It can be observed that the performance of WI-DL is better than those of other algorithms, and the classification accuracy of WI-DL improves faster than other algorithms as more samples are added. The performance of other three algorithms are ranked in the order of MUS, RS, and QBC in the experiments.

Experiments are performed on a Windows 10 computer with a 64-bit CPU intel(R) Core(TM) i5-4570s running at 2.90 GHz, and algorithms are implemented with MATLAB R2013a. Elapsed CPU times for classification of three datasets with four different algorithms are measured and summarized in Table II. It can be observed that RS is the fastest as random selection requires no computation. MUS is slower than RS but close as it only needs to compute entropy for each candidate sample. The proposed WI-DL and QBC are relatively slow. The complexity of WI-DL is mainly due to sparse coding, and the complexity of QBC is mainly due to training of different committee members. WI-DL is usually faster than QBC because a greedy-based algorithm for sparse coding is used.

TABLE V
BOTSWANA DATASETS

| Class | Label | Total | Dataset 1 | | | Dataset 2 | | | Dataset 3 | | |
|---|---|---|---|---|---|---|---|---|---|---|---|
| | | | Train (10%) | Candidate (20%) | Test (70%) | Train (20%) | Candidate (20%) | Test (60%) | Train (30%) | Candidate (20%) | Test (50%) |
| Water | 1 | 270 | 27 | 54 | 189 | 54 | 54 | 162 | 81 | 54 | 135 |
| Hippo grass | 2 | 101 | 10 | 20 | 71 | 20 | 20 | 61 | 30 | 20 | 51 |
| Floodplain grasses1 | 3 | 251 | 25 | 50 | 176 | 50 | 50 | 151 | 75 | 50 | 126 |
| Floodplain grasses2 | 4 | 215 | 22 | 43 | 150 | 43 | 43 | 129 | 65 | 43 | 107 |
| Reeds1 | 5 | 269 | 27 | 54 | 188 | 54 | 54 | 161 | 81 | 54 | 134 |
| Riparian | 6 | 269 | 27 | 54 | 188 | 54 | 54 | 161 | 81 | 54 | 134 |
| Firescar2 | 7 | 259 | 26 | 52 | 181 | 52 | 52 | 155 | 78 | 52 | 129 |
| Island interior | 8 | 203 | 20 | 41 | 142 | 41 | 41 | 121 | 61 | 41 | 101 |
| Acacia woodlands | 9 | 314 | 31 | 63 | 220 | 63 | 63 | 188 | 94 | 63 | 157 |
| Acacia shrublands | 10 | 248 | 25 | 50 | 173 | 50 | 50 | 148 | 74 | 50 | 124 |
| Acacia grasslands | 11 | 305 | 31 | 61 | 213 | 61 | 61 | 183 | 92 | 61 | 152 |
| Total | - | 2704 | 271 | 542 | 1891 | 542 | 542 | 1620 | 813 | 542 | 1349 |

### B. Experiment 2

The data used in this experiment are the airborne data from the ROSIS optical sensor, and was collected under the HySens project sponsored by the European Union. The images were acquired over the area of the University of Pavia, in northern Italy, on July 8, 2002. The number of bands of the ROSIS sensor is 115 with a spectral coverage ranging from 0.43 to 0.86 $\mu$m, and 103 hyperspectral channels are used for classification after the removal of 12 noisy bands. The spatial resolution is 1.3 m per pixel. The data have been atmospherically corrected, and the original image is shown in false color in Fig. 6(a). Fig. 6(b) shows the ground truth with a detailed view in the lower left corner. It shows ten classes in different colors, and the names of the class labels are shown on the right side.

Nine classes of interest (Asphalt, Meadows, Gravel, Trees, Metal sheets, Bare soil, Bitumen, Bricks and Shadows) have been selected for the labeled dataset. Four algorithms (WI-DL, RS, MUS, and QBC) are applied to three sets of data constructed from the ground truth data. Each set contains three classes of randomly selected data, training, candidates, and testing data, of different percentages. Table III shows the number of samples for each class of the dataset.

Fig. 7 shows results of classification done by WI-DL, RS, MUS, and QBC methods. Details of the boxed area in the middle are enlarged in the lower left corner of Fig. 7(a)–(d). The result of the proposed WI-DL algorithm shown in Fig. 7(a) matches reasonably well with the ground truth shown in Fig. 6(b). Also, the classification result of the WI-DL algorithm in Fig. 7(a) is better than results obtained by the other algorithms in Fig. 7(b)–(d), as seen in the enlarged area.

Fig. 8 shows changes of classification accuracies as number of training samples increases. It can be observed that the performance of WI-DL is better than those of other algorithms, and the classification accuracy of WI-DL improves faster than other algorithms as more samples are added. CPU times for classification of three datasets with four different algorithms are measured and summarized in Table IV. The hardware and software environments are as same as in Experiments 1.

### C. Experiment 3

The dataset used in this experiment is acquired by NASA EO-1 satellite over the Okavango Delta, in Botswana in 2001. The Hyperion sensor on EO-1 acquires data at 30-m pixel resolution over a 7.7-km strip in 242 bands covering a spectrum ranging 400–2500 nm. Preprocessing was performed by the University of Texas—Center for Space Research. Uncalibrated and noisy bands that cover water absorption features were removed, and 145 bands remained in the dataset. The data analyzed in this study, acquired on May 31, 2001, consist of observations from 14 identified classes. Fig. 9(a) shows a test image in false color, and Fig. 9(b) shows the ground truth with a detailed view in the lower left corner. It shows 12 classes in different colors, and the names of class labels are shown on the right side.

Ten classes of interest have been selected for the labeled dataset as summarized in Table V. Four algorithms (WI-DL, RS, MUS, and QBC) are applied to three sets of data constructed from the ground-truth data. Each set contains three classes of randomly selected data, training, candidates, and testing data, of different percentages. Table V shows the number of samples for each class of the dataset. The class name and the class number are given in the first and second columns, while the third column shows the total number of samples, and the rest of columns show numbers of samples in training, candidates, and test sets.

Fig. 10 shows results of classification done by WI-DL, RS, MUS, and QBC methods. Details of the boxed area in the middle is zoomed at the lower left corner of Fig. 10(a)–(d). The result of the proposed WI-DL algorithm shown in Fig. 10(a) matches reasonably well with the ground truth shown in Fig. 9(b). Also, the classification result of the WI-DL algorithm in Fig. 10(a) is better than the results obtained by other algorithms in Fig. 10(b)–(d), as seen in the enlarged area.

Fig. 11 shows changes of classification accuracies as number of training samples increases. It can be observed that the performance of WI-DL is better than those of other algorithms, and the classification accuracy of WI-DL improves faster than other algorithms as more samples are added. The performance of other three algorithms are ranked in the order of QBC, MUS, and RS in the experiments. Elapsed CPU times for classification of three datasets with four different algorithms are measured and summarized in Table VI. The hardware and software environments are as same as in Experiments 1.

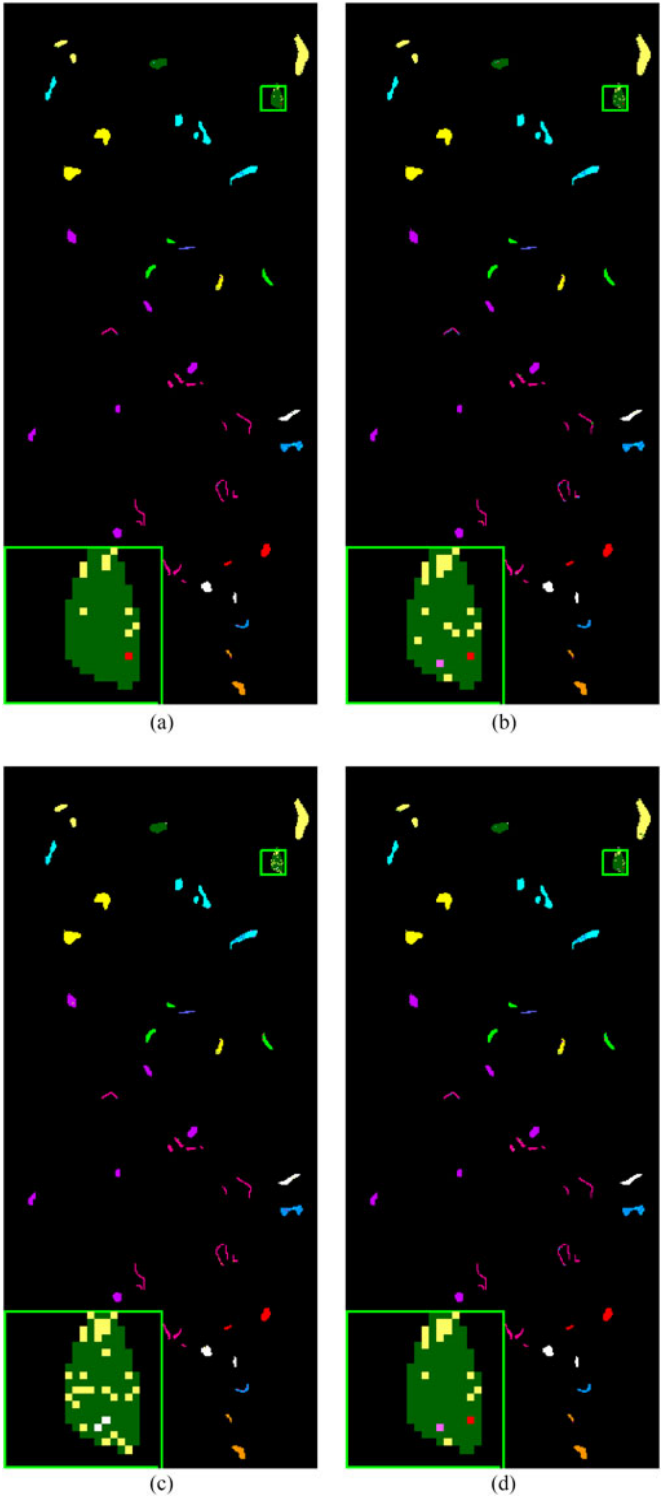

Fig. 10. Maps of classifications of different active learning methods. The accuracies for WI-DL, MUS, RS, and QBC are 91.6%, 83.4%, 76.9%, and 88.6%, respectively. (a) WI-DL, (b) MUS, (c) RS and (d) QBC.

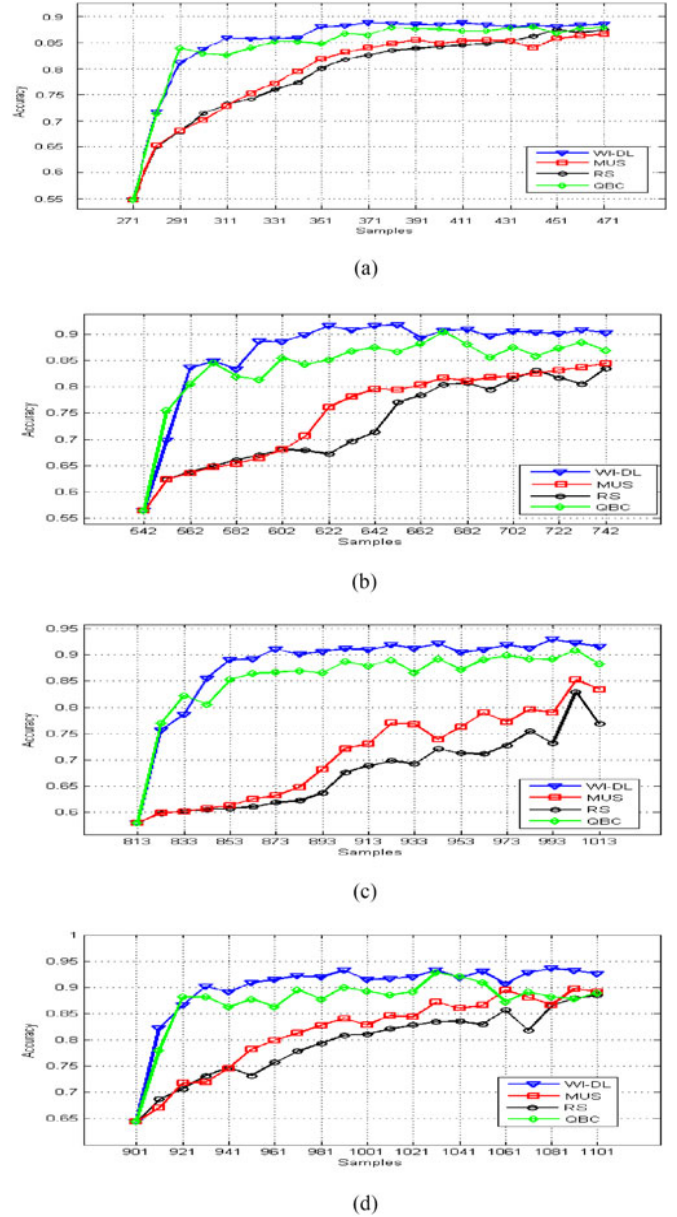

Fig. 11. Classification accuracy of different methods on different datasets. (a) Dataset 1. (b) Dataset 2. (c) Dataset 3. (d) Cross validation of 12 runs.

TABLE VI
COMPUTATION TIME (IN SECONDS) FOR BOTSWANA DATASETS

|  | WI-DL | MUS | RS | QBC |
|---|---|---|---|---|
| Dataset 1 | 43 | 22 | 20 | 47 |
| Dataset 2 | 108 | 35 | 36 | 119 |
| Dataset 3 | 127 | 51 | 52 | 149 |

## VI. CONCLUSION

In this paper, we proposed a classification algorithm based on active learning of deep networks. For active learning, additional samples to the training set are selected using the representativeness and uncertainty of the potential samples. This is achieved by integrating two criteria into a new object function. A new active learning algorithm, the WI-DL algorithm, which is suitable for searching atoms is developed by minimizing the new object function that has two criteria. The performance of the WI-DL algorithm is compared with three other methods, RS, MUS, and QBC. The proposed WI-DL performed well in the classification experiment with remotely sensed hyperspectral images. It is demonstrated that the proposed algorithm achieves higher accuracy with fewer training samples by actively selecting training samples.

ACKNOWLEDGMENT

The authors would like to express sincere gratitude to Dr. Q. Du, associate editor, and three anonymous reviewers for detailed review and constructive comments.

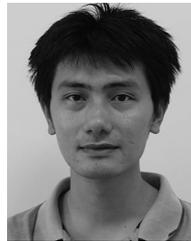

**Peng Liu** received the M.S. degree in 2004 and the Ph.D. degree in 2009, both in signal processing, from the Chinese Academic of Science, Beijing, China.

He is currently an Associate Professor at the Institute of Remote Sensing and Digital Earth, Chinese Academy of Sciences. From May 2012 to May 2013, he was with the Department of Electrical and Computer Engineering, George Washington University, Washington, DC, USA, as a Visiting Scholar. His research is focused on big data, sparse representation, compressive sensing, deep learning and their applications to remote sensing data processing.

Dr. Liu is also the Reviewer for *Applied Remote Sensing*, IEEE JOURNAL OF SELECTED TOPICS IN APPLIED EARTH OBSERVATIONS AND REMOTE SENSING, *Neurocomputing*, *Signal Processing*, etc.

**Hui Zhang** received the B.S. degree with a major in computer science from Henan Normal University in 2002, and the M.S. degree in pattern recognition and intelligent system and the Ph.D. degree in computer science, both from the Chinese Academic of Science, Beijing, China, in 2005 and 2009, respectively.

She is currently a Visiting Fellow at the Laboratory of Brain and Cognition, National Institute of Mental Health, Bethesda, MD, USA. Her current research interests include using machine learning method for big data analysis.

**Kie B. Eom** received the Ph.D. degree in electrical engineering from Purdue University, West Lafayette, IN, USA, in 1986.

He was an Assistant Professor at Syracuse University, Syracuse, NY, USA before joining the George Washington University, Washington, DC, USA, in 1989, where he is currently a Professor of electrical and computer engineering. He was also a Visiting Professor with the Center for Automation Research, University of Maryland, from 1993 to 1994. His research interests include random field modeling, automatic target recognition, speech and signal processing. His research has been sponsored by National Science Foundation, Office of Naval Research, National Security Agency, NASA, and Niagara Mohawk Power Corporation. He has consulted Enerlog Corporation, Niagara Mohawk Power Corporation, and U.S. Patent Office.

Dr. Eom is listed in Whos Who in Science and Engineering, is a senior member of IEEE Computer Society, and is a member of Tau Beta Pi, Eta Kappa Nu, and Sigma Xi.